\documentclass[11pt,a4paper]{article}
\usepackage{authblk}
\usepackage[hyperref]{aacl-ijcnlp2020}
\usepackage{times}

\usepackage{latexsym}

\usepackage{microtype}
\usepackage{multirow}
\usepackage{makecell}

\aclfinalcopy %

\title{Towards Improved Model Design for Authorship Identification:\\A Survey on Writing Style Understanding}

\author[1]{Weicheng Ma}
\author[2]{Ruibo Liu}
\author[3]{Lili Wang}
\author[4]{Soroush Vosoughi}
\affil[ ]{Minds, Machines, and Society Group}
\affil[ ] {Department of Computer Science, Dartmouth College}
\affil[1,2,3]{\texttt{\{first.last.gr\}@dartmouth.edu}}
\affil[4]{\texttt{soroush.vosoughi@dartmouth.edu}}

\date{}

\begin{document}
\maketitle
\begin{abstract}
Authorship identification tasks, which rely heavily on linguistic styles, have always been an important part of Natural Language Understanding (NLU) research.
While other tasks based on linguistic style understanding benefit from deep learning methods,
these methods have not behaved as well as traditional machine learning methods in many authorship-based tasks.
With these tasks becoming more and more challenging, however,
traditional machine learning methods based on handcrafted feature sets are already approaching their performance limits.
Thus, in order to inspire future applications of deep learning methods in authorship-based tasks in ways that benefit the extraction of stylistic features, we survey authorship-based tasks and other tasks related to writing style understanding.
We first describe our survey results on the current state of research in both sets of tasks
and summarize existing achievements and problems in authorship-related tasks.
We then describe outstanding methods in style-related tasks in general and analyze how they are used in combination in the top-performing models.
We are optimistic about the applicability of these models to authorship-based tasks and hope our survey will help advance research in this field.
\end{abstract}

\section{Introduction}
Writing style understanding is a key topic in Natural Language Processing (NLP) research.
While the content varies strongly across documents written by the same author,
stylistic features, as defined by \citet{style-hist-2}, remain constant and are the cause of a wide range of phenomena, e.g. text genre, metaphor, and irony.
An important application of stylistic features is the identification of authorship.
Because of the importance of stylistic features in NLP tasks related to authorship and other stylistic phenomena \cite{style-hist-1},
useful model architectures and feature sets are likely to be shared.
This motivates us to survey recent research achievements in style-related tasks to find possible ways of advancing the research on authorship-based tasks\footnote{Our survey covers NLU tasks only.}.

Our paper consists of three main parts.
We first introduce the NLU tasks included in our survey.
While we cannot cover all possible tasks depending on writing style understanding,
we take into account the five most-studied authorship-based tasks and nine other prevalent tasks related to writing style understanding.
Second, we provide a summary of linguistic features and neural network architectures that appear frequently in related papers.
While existing reviews have examined various solutions to these tasks, our survey emphasizes newly-emerged neural network architectures.
For example, \citet{survey-past-1} lists 21 linguistic features in five categories which are prevalent in authorship identification tasks and
\citet{survey-past-2} introduce three neural network architectures widely used in sentiment-related tasks.
We not only include these methods, but we also provide detailed analysis of techniques such as Capsule Networks (CapsNets) and Transformer networks.
In the last part, we list the models that achieve outstanding performance in authorship-based tasks and analyze their feature engineering and model designs.
Finally, to inspire the development of effective model design for authorship-based tasks, we analyze top-performing models in other style-based tasks.

The main contributions of this paper are:
\begin{itemize}
    \item %
    By expanding the scope of our review to a wider range of NLU tasks related to writing styles,
    we bridge the authorship-based tasks and other style-related tasks and enable knowledge transfer of feature engineering and model designs to authorship-based tasks.
    \item By including the most up-to-date neural network architectures in our survey,
    we demonstrate the strong possibility of applying these architectures to authorship-based tasks.
    \item By examining the detailed usage of models and/or linguistic features for extracting stylistic features,
    we provide guidance for future model design in authorship-based tasks.
\end{itemize}

\section{Tasks and Datasets}
Our survey covers a wide range of style-related tasks besides authorship-related tasks.
We provide definitions of these tasks and regularly-used standard evaluation datasets in this section.
To be consistent, in all the task definitions we use $d=\{w_1, w_2, ..., w_n\}$ to represent a document with $n$ words and $L$ to denote task-specific label sets.

\subsection{Authorship-related Tasks}
\paragraph{Authorship Attribution (AA)}
Given a document $d$ and $k$ groups of documents $D_i=\{d_1^{(i)}, d_2^{(i)}, ..., d_j^{(i)}\}$ each authored by one of $k$ authors,
the goal of AA \cite{aa-orig} is to decide whether $d$ belongs to any $D_i$ for all $0 < i \leq k$.

AA benchmark datasets include the CCAT10 and CCAT50 \cite{aa-dataset-ccat10}, IMDB62 \cite{aa-dataset-imdb62}, Blogs10 and Blogs50 \cite{aa-dataset-blogs10}, and Novel9 \cite{aa-dataset-novel9} datasets. 
Recently \citet{nlm-4} also released a Twitter-based AA dataset.
PAN shared tasks\footnote{https://pan.webis.de/shared-tasks.html} provide additional datasets for AA.
The most up-to-date in-domain AA dataset is released in PAN-12 challenges and cross-domain AA dataset in PAN-19 shared tasks.

\paragraph{Authorship Verification (AV)}
With two documents $d_1$ and $d_2$, the AV task aims at predicting whether they are written by the same author.
The Webis Authorship Verification dataset \cite{av-1} is benchmarked for AV.
PAN shared tasks provide other AV datasets in PAN-13, PAN-14, and PAN-15 challenges as well.

\paragraph{Authorship Profiling (AP)}
Given a document $d$, the AP task classifies $d$ into author groups.
There can be multiple different label sets for AP,
e.g. $L=$\{male, female\} in PAN-18 AP task and $L=$\{bot, human\_male, human\_female\} in PAN-19 AP challenge.
Celebrity Profiling (CP) is similar to AP, with the documents coming from celebrity Twitter accounts.

PAN has been hosting AP challenges from 2013, and CP since 2019. Evaluation datasets can be found in PAN shared tasks.

\paragraph{Style Change Detection (SCD)}
In a document with $k$ paragraphs $d=\{p_1, p_2, ..., p_k\}$, the goal of SCD is to detect whether any pair of adjacent paragraphs $p_i$ and $p_{i+1}$ are written by the same author or not.
SCD datasets are available in PAN challenges since the year of 2017.

\subsection{Other Style-Related Tasks}
\paragraph{Sentiment Analysis (SA)}
Given a document $d$, the aim of SA \cite{sa-orig} is to predict the sentiment class
or sentiment polarity associated with $d$. SA labels can be either from a discrete
label set $L=$ \{Negative[, Neutral], Positive\} or a sentiment polarity score in $L = [-1, 1]$.

Stanford Sentiment Treebank \cite{sst-orig} is a standard evaluation bed for SA.
The RT dataset \cite{rt-orig} is constructed from Internet movie reviews and it provides two types of sentiment labels.
Yelp\footnote{https://www.yelp.com/dataset} also provides a large-scale SA dataset based on business reviews.

\paragraph{Aspect-Based Sentiment Analysis (ABSA)}
Different from SA, ABSA \cite{absa-orig} evaluates the sentiment of $d$ on each aspect $a$ in $d$. 
Each aspect $a=\{w^a_1, w^a_2, ..., w^a_k\}$ is either a substring in $d$ or an aspect type (e.g. restaurant).
There can be multiple aspects in $d$ and the sentiment polarity on each aspect does not have
to be towards the same direction \cite{absa-review}. 
The International Workshop on Semantic Evaluation (SemEval) provides multiple ABSA evaluation datasets \cite{absa-dataset-1,absa-dataset-2,absa-dataset-3,absa-dataset-4}.
\citet{absa-dataset-5} also release an ABSA dataset based on Twitter posts.

\paragraph{Stance Detection (SD)}
SD \cite{sd-orig-1,sd-orig-2} takes two inputs, a document $d$ and a target $c$. 
$c$ can be either an entity (e.g. a person) or a claim. 
The goal is to classify the stance of $d$ towards $c$ into the label set $L=$ \{Agree, Disagree[, Discuss, Unrelated]\}. 
$c$ does not have to physically appear in $d$. 
The number of $c$ per $d$ is not limited either.

One benchmark dataset for SD is from SemEval-2016 Task 6 \cite{sd-dataset-1}. 
The Brexit Blog Corpus \cite{sd-dataset-2}, US Election Tweets Corpus \cite{sd-dataset-3}, and Moral Foundations Twitter Corpus \cite{sd-dataset-4} have also received much attention recently. 
\citet{sd-dataset-5} publish an SD dataset where each document contains two targets and two stance labels.
Some other related datasets are provided in the review by \citet{sd-review}.

\paragraph{Emotion Recognition (ER)}
ER is closely related to SA, but with more fine-grained labels and better expressiveness \cite{er-sa}. 
ER classifies $d$ into $L$ containing pre-defined emotion types or draws intensity scores of $d$ over 
the emotion types in $L$. 
The most common ER label set contains eight basic emotion types, i.e. $L=$ \{Joy, Sadness, Anger, 
Fear, Anticipation, Surprise, Love, Disgust, Neutral\} \cite{er-types}. 

The standard evaluation datasets of uni-modal ER consist of the EmoBank \cite{er-dataset-emobank}
and MELD \cite{er-dataset-meld} datasets. Benchmark datasets on multi-modal ER include the IEMOCAP \cite{er-dataset-iemocap}, MOUD \cite{er-dataset-moud}, 
ICT-MMMO \cite{er-dataset-ictmmo}, MOSI \cite{er-dataset-mosi} and MOSEI \cite{er-dataset-mosei} datasets.

\paragraph{Metaphor Detection (MD)}
A word or phrase is metaphorical if it bears different meaning from its literal semantic meaning in a context. 
MD is designed to detect whether $d$ or any word group $w_i,...,w_{i+m}$ in $d$ is metaphorical.
Benchmarked datasets on MD include MOH \cite{md-dataset-moh} and its subset MOH-X, TSV \cite{md-dataset-tsv}, TroFi \cite{md-dataset-TroFi}, VUA \cite{md-dataset-vua}, and LCC \cite{md-dataset-lcc}.

\paragraph{Other Tasks}
The Irony Detection (ID), Offense Detection (OD), Formality Classification (FC), and Humor Detection (HD) tasks are all document-level classification tasks.
ID detects whether the contextual meaning of $d$ is opposite to its literal meaning.
SARC \cite{id-dataset-1} is a common evaluation dataset on ID.
\citet{id-dataset-2} also release a multi-modal ID dataset named MUStARD.
OD classifies $d$ into $L=$\{Offensive, Non-offensive\}.
Subtasks of OD include the detection of hate speech, cyber-bulling and cyber-aggression \cite{od-dataset-1}.
\citet{od-dataset-1} and \citet{od-dataset-2} provide two benchmarked OD datasets.
FC evaluates whether $d$ is formal or informal and HD identifies humorous writing styles from $d$.
\citet{fc-dataset-1} provide a large-scale dataset for formality transfer, while its labels fit the FC task perfectly.
\citet{transformer-1} label a HD dataset based on data from Reddit, Kaggle, and Pun of the Day.
\citet{hd-dataset-1} make HD more fine-grained by introducing eight humor categories and five humor levels to the annotations.

\section{Methods}
In this section, we summarize useful features and common neural network architectures in style-related tasks.
We also display the various feature engineering methods and model designs to inspire future research on writing style understanding.

\subsection{Linguistic Features}
\paragraph{Language Model Features}
N-Gram features on both word and character levels are among the most frequently used language model features in representing writing styles.
The most common choices of n-gram features are word-level 1-, 2-, and 3-gram and character-level 3-, 4-, and 5-gram features.
\citet{ngram-2} additionally show the importance of punctuation marks in the AA task using 
an n-gram model with everything except for punctuation marks masked by asterisk symbols (*).
\citet{ngram-5} claim that digits and named entities are also key identifiers of writing styles.
\citet{ngram-1} attribute the advantage of using character-level n-gram features to the high priority of subword features (e.g. suffixes and prefixes) in authorship-related tasks.
\citet{ngram-4} loosen the constraint of n-gram features and consider the co-occurrence of word pairs instead.
Statistical features, e.g. TF-IDF scores, are regularly used in combination to the language-model-based features to avoid overemphasizing stop words \cite{ngram-3}.
As text representations generated from n-gram features tend to be high-dimensional and sparse,
\citet{pca-1} apply Principal Component Analysis to compress them into low-dimensional vectors.
\citet{er-sa} achieve the similar goal using topic models (e.g., Latent Dirichlet Allocation).
Recently, many researchers have turned to neural language models (e.g. Skip-gram model, \citet{skipgram-orig}) for text representations \cite{nlm-1,nlm-2,nlm-3}.
Word vectors from neural language models can be compared, clustered, and used in any statistical method to form document representations \cite{nlm-4,nlm-5}.

\paragraph{Syntactic Features}
Part-of-Speech (POS) tags and dependency relations are two main sets of syntactic features used in style-related tasks.
The coexistence of a set of POS forms syntactic patterns that can be used to recognize stylistic phenomena, e.g. complaints \cite{pos-2}.
\citet{pos-1} also argue that the pattern of using each POS in a document is related to the formality of the text.
Dependency parsing is in most cases used in combination to POS tagging to help extract meaningful syntactic patterns.
\citet{parsing-1} claim that pure POS tags are simplistic and are only able to model shallow syntactic features, while dependency parsing provides richer authorship information.
Their experiments suggest that jointly using POS and dependency features results in the best performance.
The reason why syntactic features are important in style-based tasks is clear:
writing styles are usually specific to the choice and arrangement of words in a document,
so the interrelationship among words is crucial.

\paragraph{Lexical Features}
When addressing stylistic phenomena including sentiments and emotions, 
specifically-designed lexicons provide solid help, especially to short and simple documents.
\citet{pos-2}, for example, approach the SA task with the help of MPQA \cite{lexicon-1} lexicon for sentiment, and the NRC \cite{lexicon-2.1,lexicon-2.2} and Volkova \& Bachrach \cite{lexicon-3} lexicons for sentiment and emotion.
The function of lexicons, however, is limited as stylistic phenomena are often caused by linguistic expressions in a long span of text.
According to \citet{pos-2}, the best-performing lexicon in their experiments produces worse results than a simple bag-of-words model.
The semantic relatedness between words is beneficial to style-related tasks as well.
WordNet \cite{wordnet} is the most heavily used lexical database of semantic relation information among words, according to our survey.
\citet{lexicon-4} use hypernyms and synonyms of words from WordNet to disambiguate them when discovering nuanced metaphors from text.
Since different authors tend to express different attitudes towards specific events,
lexical features can help solve simple cases in authorship-related tasks with high accuracy and thus reduce noise when training complex models.

\paragraph{Other Features}
In our survey, we came across some linguistic features that are used for specific tasks.
We group these features here since we think they can be potentially help detect authorship information.
Downgraders and politeness are used by \citet{pos-2} to identify complaints and \citet{task-specific-1} introduce alliterations, slangs, and rhyming to the HD task.
These features are all author-specific language-use features and qualify for coarse-grained authorship identification.
Hashtags, URLs, retweets, tweet popularity, elongated words, hapax legomena, and superlatives are valuable features for AA \cite{nlm-4},
and they well fit all the authorship-based tasks in the social media domain, e.g. the CP task.

\subsection{Neural Network Architectures}
\paragraph{Convolutional Neural Networks (CNNs)}
CNNs \cite{cnn-orig} extract local dependency features from positionally nearby tokens, 
which act similarly to n-gram models with n bounded by the sizes of convolution kernels.
Benefited from parameter sharing, the training process of CNNs is easy and time-efficient.
\citet{cnn-1} combine the features extracted by multiple one-layer CNNs, each with different kernel sizes,
into document representations.
While CNN-based models on style-related tasks generally share the same architecture,
their inputs and external resources usually vary.
\citet{cnn-2} apply CNNs on text embeddings explicitly enriched by n-gram and syntactic features.
They show by experiments that these additional features help their model achieve better performance than vanilla CNNs.
\citet{cnn-3} argue that character-level CNNs are superior in capturing stylistic features.
This attributes to the ability of character-level CNNs in modeling sub-word features, e.g. prefixes,
which are important style markers.
Similarly, \citet{cnn-4} convert documents into character-level bigram embeddings and apply CNN feature extractors on these feature maps.
Lexical features can also be used in conjunction with CNNs.
For example, \citet{cnn-5} keep pre-trained sentiment embeddings of words from multiple domains in a memory module and inject the lexical information into CNN encodings of documents by appending normal word vectors attended by the sentiment embeddings to the end of the CNN output.

\paragraph{Graph Neural Networks (GNNs)}
GNNs have recently attracted a lot of research attentions in the NLP community.
Popular graph models on style-related tasks include Graph Convolutional Networks (GCNs) \cite{gcn-orig} and Graph Attention Networks (GATs) \cite{gat-orig}.
While GCNs are similar to CNNs, they convolve on graphs, treating nodes connected by edges as neighbors regardless of the absolute positions of words.
In addition to connectivity, GATs generate the representation of each node by applying self-attention mechanism on all its neighbor nodes.
These characteristics make GNNs appropriate for encoding dependency information.
Researchers have been using multi-layer GCNs \cite{gcn-1.1,gcn-1.2} and multi-layer GATs \cite{gat-1} to encode documents after syntactic parsing for the ABSA task, for example.
In these scenes, GCNs and GATs are usually used jointly with other feature extractors (e.g. a Long-Short-Term Memory (LSTM) model) since syntactic features solely are not enough for style-related tasks.
From this perspective, GCNs and GATs are valuable model architectures in authorship-related tasks where syntactic information is crucial.
As \citet{gcn-2} additionally use GCN to capture high-level features across dialogue utterances,
we can foresee the application of GNNs in SCD tasks by viewing the paragraphs as dialogue utterances.

\paragraph{Recurrent Neural Networks (RNNs)}
The use of RNNs is widespread in NLP tasks since RNNs are designed to encode sequential data, 
and because they are able to preserve long-term dependency information.
LSTM and Gated Recurrent Units (GRU) are the two most prevalent model architectures in the family of RNNs.
\citet{rnn-1} show an example of approaching the SA task with a one-layer uni-directional GRU model.
Similar to CNNs, RNN-based models are also used to extract character-level dependency features in style-related tasks \cite{rnn-6}.
Multiple RNN-based model architectures have been invented to improve their encoding ability.
Hierarchical RNNs \cite{hiernn} are designed to extract features at different abstraction levels.
\citet{rnn-2} break documents down to word- and sentence-level pieces and use two LSTM models to generate sentence- and document-level encodings, respectively.
Stacked RNNs are relatively rare in our survey.
\citet{rnn-3} examine two-layer LSTM and GRU models on the SA task and their performances are very closed to a single-layer CNN model.
Rather, bidirectional RNNs (BiRNNs, \citet{birnn}) or stacked BiRNNs are popular for their ability to use information flows from both directions of the input sequence, which helps with a lot of style-related tasks 
\cite{rnn-4.1,rnn-4.2,rnn-4.3,rnn-4.4}.
Attention mechanism \cite{attention-cho} is applicable to all the RNN architectures we describe above.
In pure RNN models, each token contributes to the output embeddings equally,
while some words are in fact more important to the predictions than the rest.
Impaired verbs and objects in a sentence are strong signs of metaphors, for example.
Attention mechanism teaches the RNN-based models to dynamically weigh the token embeddings and thus improves their performances \cite{rnn-5.1,rnn-5.2,rnn-5.3}.
In recent research, RNNs are frequently used in combination to CNNs as well.
There are two main forms of assembling CNNs and RNNs,
one of which is separately encoding the document with RNN and CNN models and concatenating the output vectors together \cite{rnn-cnn-1} or combining the predictions made by both models through majority voting \cite{rnn-cnn-2}.
The second form is to stack the CNN and RNN models together and have the CNN model extracting local dependency features on the output of the RNN model or the other way around \cite{rnn-cnn-3}.

\paragraph{Memory Networks (MemNNs)}
MemNNs \cite{memnn-orig} encode sequences with the help of external memories storing global information and preceding states.
Initialization, read, and update are three basic operations to the memory cells in MemNNs,
the latter two of which are usually implemented using attention mechanism.
Multi-hop memories are used dominantly in style-related tasks as they preserve features from multiple different abstraction levels \cite{dl-lecun}.
\citet{memnn-1} apply MemNNs in an ABSA task.
Their model stacks embeddings of context words to initialize external memory and queries the memory with embeddings of the aspect terms.
In recent research, MemNNs are usually used in parallel with other neural networks, e.g. RNNs and CNNs,
to keep richer contextual information in the memories.
\citet{memnn-rnn-1} initialize and update the memories both with the help of GRU,
while in \citet{memnn-rnn-cnn-1} the memory cells are initialized by an ensembled CNN-GRU feature extractor.
Memory cells in a MemNN model can save author-specific information so we are especially optimistic about its use in AA and AP tasks.

\begin{table*}[h!]
\centering
\begin{tabular}{llcccc}
\Xhline{2\arrayrulewidth}
\multicolumn{2}{c}{Models}                        & \multicolumn{4}{c}{Datasets}                                       \\ \Xhline{2\arrayrulewidth}
\multicolumn{2}{c}{AA}                            & CCAT10         & CCAT50         & IMDB62         & PAN-12          \\ \Xhline{2\arrayrulewidth}
\citet{ngram-5}                  & CN-gram+SVM        & 79.60          & -              & -              & -               \\ \hline
\citet{aa-2}                  & CN-gram+FastText   & 74.80          & 72.60          & 94.80          & -               \\ \hline
\multirow{2}{*}{\citet{cnn-2}} & CN-gram+CNN        & 86.80          & 76.50          & 95.21          & -               \\ \cline{2-6} 
                             & CN-gram+Syntax+CNN & \textbf{88.20} & \textbf{81.00} & \textbf{96.16} & -               \\ \hline
\citet{aa-4}                  & Syntax+CNN+RNN     & -              & -              & -              & \textbf{100.00} \\ \hline
\citet{aa-5}                  & SVM                & -              & -              & -              & 85.71           \\ \Xhline{2\arrayrulewidth}
\end{tabular}
\caption{Model architectures and accuracy scores on four mainstream AA datasets. CN-gram refers to character-level n-gram. PAN-12 displays accuracy scores on task I of the 2012 PAN Authorship Attribution shared task. The best score on each dataset is in bold.}
\label{tbl:aa-results}
\end{table*}

\paragraph{Capsule Networks (CapsNets)}
CapsNets \cite{capsnet-orig} feature spatial relation modeling and the dynamic routing mechanism. 
CapsNets internally cluster the encodings of similar entities (e.g. words with the same sentiment labels) together in a capsule and adjust the encodings' attributes (e.g. sentiment polarity) with affine operations. 
The dynamic routing mechanism enables CapsNets to reduce the effect of noise words on the representation of the entire text. 
These features make CapsNets sensitive to writing styles and robust to irrelevant information.
\citet{capsnet-1} use dynamic routing to select appropriate n-gram groups for predictions in the ABSA task.
\citet{capsnet-2} show by experiments that even with simple dynamic routing strategy (e.g. using averaged embeddings of sentiment words), CapsNets are able to boost the performance of complex-enough models (e.g. BERT \cite{bert-orig}, a pre-trained Transformer-based model).
\citet{capsnet-3} utilize CapsNets to address zero-shot learning problem.
All the above examples show that CapsNets are efficient in extracting and clustering style markers, and making predictions accordingly.
So the use of CapsNets solely or combined with other neural networks on authorship-based tasks is promising and deserves more research attention.

\paragraph{Transformer Networks}
Transformer Networks \cite{transformer-orig} is a newly-emerged family of neural networks,
the core of which are self-attention and positional encoding.
Models based on Transformer Networks are usually deep, with giant amounts of parameters.
In most cases, Transformer-based models are pre-trained on large unlabeled corpora and fine-tuned on task-specific datasets, e.g. BERT.
Benefited from their large sizes of parameters and training data,
Pre-trained Transformer-based models have been refreshing the records on style-related tasks since their emergence.
For example, \citet{transformer-1} use BERT to encode text for the HD task and their model achieves results comparable to human annotators without feature engineering or model restructuring.
\citet{transformer-2} use external knowledge from ConceptNet \cite{conceptnet-orig} and the NRC\_VAD lexicon \cite{nrc-orig} to augment word embeddings with emotion information,
and they introduce dialogue structure information to the self-attention mechanism in Transformer networks.
This shows the advantage of combining Transformer-based models with handcrafted feature sets or other neural network architectures,
providing possibilities of improving the performance of Transformer networks in authorship-based tasks where handcrafted feature sets are extensively researched.

\section{Discussions}
In this section, we first explain in more details the feature engineering methods and model designs that achieve good results in extracting authorship information.
We also analyze the causes that some methods do not behave as well as expected.
Next, we do similar analysis for the SA, ABSA, and ER tasks to infer if any specific model design can be transferred to authorship-based tasks.
These three tasks are among the most well-developed style-based tasks and the models we discuss are also top-performing models in these tasks.

We show the best-performing models with different classifier architectures on four AA datasets in Table \ref{tbl:aa-results} to discuss current research on the AA task.
We choose PAN-12 instead of PAN-19 since the PAN-12 AA task is under in-domain settings, similar to other tasks we list.
According to the PAN official reports \cite{pan-19-report,pan-12-report},
existing systems are based on similar feature sets and classifiers so the task choice does not affect our analysis.
From the table we can see the importance of character-level n-gram features for the AA task.
Syntactic features additionally show great potentials in this task, according to our observations.
As for the choice of model architectures, a shallow neural model (e.g. FastText) does not beat SVM classifiers with similar feature sets,
while CNNs show superior encoding ability on the AA task.
The combination of CNNs and RNNs with POS features also outperform the SVM model on the PAN-12 Task I, achieving 100\% accuracy on the test set.
The task, however, is too constrained in data size, 
with 28 training examples and 14 test examples \cite{pan-12-report}.
As a result, models with handcrafted features and probabilistic models also perform well.
Evaluations of the model by \citet{aa-4} on more complex AA tasks should be performed to show its strength and robustness.
Existing models on the PAN AV challenges almost entirely rely on handcrafted features sets.
Character-level n-gram features are again the most favored linguistic features.
\citet{av-1} construct character trigram vectors for the documents and evaluate the differences between each pair of documents as features using seven distance measures.
\citet{rnn-6} uses an RNN-based model on character level to verify authorship and achieves higher scores than \citet{av-1} on PAN-15 AV task,
proving the power of deep neural networks on authorship-based tasks.

\begin{table}[]
\centering
\begin{tabular}{lcc}
\Xhline{2\arrayrulewidth}
\multicolumn{1}{c}{Models} & \multicolumn{2}{c}{Subtasks}                                      \\ \Xhline{2\arrayrulewidth}
\multicolumn{1}{c}{AP}     & Bot                             & Gender                          \\ \Xhline{2\arrayrulewidth}
\citet{ap-1}               & \multirow{2}{*}{\textbf{95.95}} & \multirow{2}{*}{83.79}          \\
Random Forest                         &                                 &                                 \\ \hline
\citet{ap-2}               & \multirow{2}{*}{90.61}          & \multirow{2}{*}{\textbf{84.32}} \\
Logistic Regression                         &                                 &                                 \\ \hline
\citet{ap-3}               & \multirow{2}{*}{93.33}          & \multirow{2}{*}{83.60}          \\
Transformer                       &                                 &                                 \\ \hline
\citet{ap-4}               & \multirow{2}{*}{91.36}          & \multirow{2}{*}{75.72}          \\
RNN                        &                                 &                                 \\ \hline
\citet{ap-5}               & \multirow{2}{*}{91.82}          & \multirow{2}{*}{79.73}          \\
CNN                        &                                 &                                 \\ \hline
\citet{ap-6}               & \multirow{2}{*}{90.08}          & \multirow{2}{*}{77.58}          \\
CNN+RNN                    &                                 &                                 \\ \Xhline{2\arrayrulewidth}
\end{tabular}
\caption{Accuracy scores on PAN-19 AP challenge. Bot refers to the Bot/user detection subtask and Gender is the author gender prediction subtask of the challenge.}
\label{tbl:ap-results}
\end{table}
\begin{table*}[h!]
\centering
\begin{tabular}{llcccc}
\Xhline{2\arrayrulewidth}
\multicolumn{2}{c}{Models}                          & \multicolumn{4}{c}{Datasets}                            \\ \Xhline{2\arrayrulewidth}
\multicolumn{2}{c}{SA}                              & SST-2        & SST-5        & RT           & Yelp       \\ \Xhline{2\arrayrulewidth}
\multirow{3}{*}{\citet{sa-1}} & CNN                 & 81.60        & 41.99        & 76.12        & 61.18      \\ \cline{2-6} 
                              & RNN                 & 81.58        & 41.67        & 76.22        & 61.86      \\ \cline{2-6} 
                              & Transformer         & \textbf{93.03}        & \textbf{55.38}        & \textbf{88.68}        & \textbf{67.85}      \\ \Xhline{2\arrayrulewidth}
\multicolumn{2}{c}{ABSA}                            & Twitter      & Laptop14     & Rest14       & Rest16     \\ \Xhline{2\arrayrulewidth}
\citet{capsnet-2}               & CapsNets+Transformer   &     -    &    -      & \textbf{85.93}        &    -    \\ \hline
\citet{gcn-1.1}                & RNN+GCN             & 73.66        & 72.99        & 74.02       & \textbf{69.93}      \\ \hline
\citet{absa-2}                & RNN+CNN+Attention   & \textbf{77.72}        & \textbf{73.84}        & 72.90        &      -      \\ \hline
\citet{gat-1}            & RNN+Transformer+GAT &       -       & 80.10        & 83.00        &      -      \\ \Xhline{2\arrayrulewidth}
\multicolumn{2}{c}{ER}                              & \multicolumn{2}{c}{IEMOCAP} & \multicolumn{2}{c}{MELD}  \\ \Xhline{2\arrayrulewidth}
\citet{memnn-rnn-cnn-1}       & CNN+RNN+MemNN       & \multicolumn{2}{c}{63.50}   & \multicolumn{2}{c}{-}      \\ \hline
\citet{gcn-2}                 & RNN+Attention+GCN   & \multicolumn{2}{c}{\textbf{64.18}}   & \multicolumn{2}{c}{58.10} \\ \hline
\citet{transformer-2}         & Transformer         & \multicolumn{2}{c}{59.56}   & \multicolumn{2}{c}{\textbf{58.18}} \\ \Xhline{2\arrayrulewidth}
\end{tabular}
\caption{Evaluation results on the SA, ABSA, and ER tasks. SST-2 and SST-5 refer to the binary and fine-grained Stanford Sentiment Treebank datasets, respectively; Laptop14 and Rest14 are the laptop and restaurant datasets from SemEval 2014 Task 4; Rest16 is from SemEval 2016 Task 5. Accuracy scores are evaluated for SST-2, SST-5, RT, and Yelp challenge datasets and F-1 scores for the rest tasks.}
\label{tbl:style-results}
\end{table*}

Table \ref{tbl:ap-results} lists the performances of six systems, each with different model architectures, on the PAN-19 AP shared task \cite{pan-19-ap-orig}.
It is worth noting that deep neural models perform substantially worse than the Random Forest and Logistic Regression models.
This is most likely due to the limited training data size (2,060 records for bot/user specification and 2,060 for gender prediction subtasks).
The BERT-based model \cite{ap-3} generates better results than the RNN- or CNN-based models probably because it has been pre-trained on a large corpus.
In the AP task, word- and character-level n-gram features are still among the top choices and are used in all the models listed in Table \ref{tbl:ap-results}.
For example, \citet{ap-1} bases their predictions on the TF-IDF scores of 300 most frequent word unigrams in the training set.
On the other hand, since the corpus is made up of tweets,
almost all the systems consider tweet-specific linguistic features including the counts of URLs, retweets, mentions, emojis, wrongly-spelled words, hashtags, etc.
\citet{ap-2} and \citet{ap-6} replace emojis with their respective descriptive phrases at the preprocessing stage to provide the models with enriched stylistic information.
All the knowledge learned from the AP tasks well apply to the PAN-19 CP challenge \cite{cp-orig} where the number of instances in each class is very imbalanced, e.g., \citet{cp-1} show that directly using BERT or other deep neural networks is not as competitive as statistical or probabilistic methods.

The design of neural models has been the major topic in a wide range of style-based tasks,
while neural networks are mostly used in the simplest way in authorship-related tasks.
To better fit deep neural networks to authorship-related tasks,
we also review the usage of neural models in other style-related tasks.
Table \ref{tbl:style-results} displays current top-ranking results and their model architectures on SA, ABSA, and ER tasks.
Directly applying pre-trained Transformer-based models on the text generates good results except when the utterances are short and transitions between them are frequent \cite{transformer-2}.
This suggests that Transformer Networks can potentially be applied to AA and AV tasks without much modification, if the training instances in each class are enough to fine-tune a Transformer-based model.
Using Transformer-based models in the SCD task is also promising since they are stronger than RNNs and CNNs in encoding long text pieces, e.g. paragraphs, in an article.
CapsNets are effective extractors for stylistic features that can probably be used to solve authorship-based tasks since their dynamic routing mechanism benefits the identification of style markers.
\citet{capsnet-2} demonstrate the power of CapsNets by the combined use of CapsNets and Transformer networks in an ABSA task.
\citet{sa-1} additionally introduce adversarial training \cite{adversarial-orig} to train a robust classifier.
Similar methods can be used in authorship-based tasks, e.g. the AP and CP tasks, to compensate for the lack of training data in small classes, relieving the problem of data imbalance.

It is also noteworthy that attention mechanism is widely used in the style-related tasks we surveyed,
while participants in the authorship identification tasks seldom take advantage of attentions.
The attention mechanism weighs each token in the text differently according to their interrelationships,
in order not to confuse the classifier with unimportant tokens.
This can be implanted to authorship-related tasks as well since style markers are sparse compared to other words, and as useful information for classification can often be located easily even using handcrafted features only.
The combined use of CNNs and RNNs or MemNNs also show strong ability in style-related tasks,
as is consistent with results shown in Table \ref{tbl:aa-results}.
Besides hard-coding POS or dependency relationship information into the inputs of CNNs,
GNNs used in combination with dependency features show strong ability on the ABSA and ER tasks, implying the importance of syntactic features.
These light-weight neural network architectures, though not beating the performances of Transformer-based models on many tasks, are probably more suitable for authorship-related shared tasks in PAN since they require much less data to train.
They can also be used in combination to the handcrafted feature sets easily, without the need for restructuring the models and retraining the weights.

\section{Conclusion}
The identification of authorship is a foundational task in NLP research.
In this paper, we analyzed linguistic features and model architectures in recent authorship-based research.
Our survey indicates that though feature engineering has been studied extensively for this task,
research on the use of neural network models is not as up-to-date as in other style-related tasks.
As authorship-based tasks are becoming increasingly complex, the sole use of traditional machine learning models with handcrafted feature sets will be less competitive.
To better take advantage of the power of deep neural networks, we examined the model design in various writing style understanding tasks which we hope will inspire future research on authorship-based tasks.

\bibliography{anthology}

\begin{thebibliography}{131}
\expandafter\ifx\csname natexlab\endcsname\relax\def\natexlab#1{#1}\fi

\bibitem[{Akhtar et~al.(2017)Akhtar, Kumar, Ghosal, Ekbal, and
  Bhattacharyya}]{rnn-3}
Md~Shad Akhtar, Abhishek Kumar, Deepanway Ghosal, Asif Ekbal, and Pushpak
  Bhattacharyya.
\newblock A multilayer perceptron based ensemble technique for fine-grained
  financial sentiment analysis.
\newblock In \emph{Proceedings of the 2017 Conference on Empirical Methods in
  Natural Language Processing}, pages 540--546, Copenhagen, Denmark.
  Association for Computational Linguistics.
\newblock

\bibitem[{Akiva(2012)}]{aa-5}
Navot Akiva.
\newblock Authorship and plagiarism detection using binary bow features.
\newblock In \emph{CLEF (Online Working Notes/Labs/Workshop)}.

\bibitem[{Arag{\'o}n et~al.(2019)Arag{\'o}n, L{\'o}pez-Monroy,
  Gonz{\'a}lez-Gurrola, and Montes-y G{\'o}mez}]{nlm-5}
Mario~Ezra Arag{\'o}n, Adrian~Pastor L{\'o}pez-Monroy, Luis~Carlos
  Gonz{\'a}lez-Gurrola, and Manuel Montes-y G{\'o}mez.
\newblock Detecting depression in social media using fine-grained emotions.
\newblock In \emph{Proceedings of the 2019 Conference of the North {A}merican
  Chapter of the Association for Computational Linguistics: Human Language
  Technologies, Volume 1 (Long and Short Papers)}, pages 1481--1486,
  Minneapolis, Minnesota. Association for Computational Linguistics.
\newblock

\bibitem[{Bagher~Zadeh et~al.(2018)Bagher~Zadeh, Liang, Poria, Cambria, and
  Morency}]{er-dataset-mosei}
AmirAli Bagher~Zadeh, Paul~Pu Liang, Soujanya Poria, Erik Cambria, and
  Louis-Philippe Morency.
\newblock Multimodal language analysis in the wild: {CMU}-{MOSEI} dataset and
  interpretable dynamic fusion graph.
\newblock In \emph{Proceedings of the 56th Annual Meeting of the Association
  for Computational Linguistics (Volume 1: Long Papers)}, pages 2236--2246,
  Melbourne, Australia. Association for Computational Linguistics.
\newblock

\bibitem[{Bagnall(2015)}]{rnn-6}
Douglas Bagnall.
\newblock Author identification using multi-headed recurrent neural networks.
\newblock In \emph{Working Notes of {CLEF} 2015 - Conference and Labs of the
  Evaluation forum, Toulouse, France, September 8-11, 2015}, volume 1391 of
  \emph{{CEUR} Workshop Proceedings}. CEUR-WS.org.
\newblock

\bibitem[{Bahdanau et~al.(2015)Bahdanau, Cho, and Bengio}]{attention-cho}
Dzmitry Bahdanau, Kyunghyun Cho, and Yoshua Bengio.
\newblock Neural machine translation by jointly learning to align and
  translate.
\newblock In \emph{3rd International Conference on Learning Representations,
  {ICLR} 2015, San Diego, CA, USA, May 7-9, 2015, Conference Track
  Proceedings}.
\newblock

\bibitem[{Basile et~al.(2019)Basile, Gatt, and Nissim}]{parsing-1}
Angelo Basile, Albert Gatt, and Malvina Nissim.
\newblock You write like you eat: Stylistic variation as a predictor of social
  stratification.
\newblock In \emph{Proceedings of the 57th Annual Meeting of the Association
  for Computational Linguistics}, pages 2583--2593, Florence, Italy.
  Association for Computational Linguistics.
\newblock

\bibitem[{Bevendorff et~al.(2019)Bevendorff, Hagen, Stein, and Potthast}]{av-1}
Janek Bevendorff, Matthias Hagen, Benno Stein, and Martin Potthast.
\newblock Bias analysis and mitigation in the evaluation of authorship
  verification.
\newblock In \emph{Proceedings of the 57th Annual Meeting of the Association
  for Computational Linguistics}, pages 6301--6306, Florence, Italy.
  Association for Computational Linguistics.
\newblock

\bibitem[{Birke and Sarkar(2006)}]{md-dataset-TroFi}
Julia Birke and Anoop Sarkar.
\newblock A clustering approach for nearly unsupervised recognition of
  nonliteral language.
\newblock In \emph{11th Conference of the {E}uropean Chapter of the Association
  for Computational Linguistics}, Trento, Italy. Association for Computational
  Linguistics.
\newblock

\bibitem[{Biswas et~al.(2015)Biswas, Chadda, and Ahmad}]{rnn-1}
Shamim Biswas, Ekamber Chadda, and Faiyaz Ahmad.
\newblock Sentiment analysis with gated recurrent units.
\newblock \emph{Department of Computer Engineering. Annual Report Jamia Millia
  Islamia New Delhi, India}.

\bibitem[{Bolonyai et~al.(2019)Bolonyai, Buda, and Katona}]{ap-4}
Flora Bolonyai, Jakab Buda, and Eszter Katona.
\newblock Bot or not: {A} two-level approach in author profiling.
\newblock In \emph{Working Notes of {CLEF} 2019 - Conference and Labs of the
  Evaluation Forum, Lugano, Switzerland, September 9-12, 2019}, volume 2380 of
  \emph{{CEUR} Workshop Proceedings}. CEUR-WS.org.
\newblock

\bibitem[{Buechel and Hahn(2017)}]{er-dataset-emobank}
Sven Buechel and Udo Hahn.
\newblock {E}mo{B}ank: Studying the impact of annotation perspective and
  representation format on dimensional emotion analysis.
\newblock In \emph{Proceedings of the 15th Conference of the {E}uropean Chapter
  of the Association for Computational Linguistics: Volume 2, Short Papers},
  pages 578--585, Valencia, Spain. Association for Computational Linguistics.
\newblock

\bibitem[{Busso et~al.(2008)Busso, Bulut, Lee, Kazemzadeh, Mower, Kim, Chang,
  Lee, and Narayanan}]{er-dataset-iemocap}
Carlos Busso, Murtaza Bulut, Chi-Chun Lee, Abe Kazemzadeh, Emily Mower, Samuel
  Kim, Jeannette~N. Chang, Sungbok Lee, and Shrikanth~S. Narayanan.
\newblock Iemocap: interactive emotional dyadic motion capture database.
\newblock \emph{Language Resources and Evaluation}, 42(4):335.
\newblock

\bibitem[{Castro et~al.(2019)Castro, Hazarika, P{\'e}rez-Rosas, Zimmermann,
  Mihalcea, and Poria}]{id-dataset-2}
Santiago Castro, Devamanyu Hazarika, Ver{\'o}nica P{\'e}rez-Rosas, Roger
  Zimmermann, Rada Mihalcea, and Soujanya Poria.
\newblock Towards multimodal sarcasm detection (an {\_}{O}bviously{\_} perfect
  paper).
\newblock In \emph{Proceedings of the 57th Annual Meeting of the Association
  for Computational Linguistics}, pages 4619--4629, Florence, Italy.
  Association for Computational Linguistics.
\newblock

\bibitem[{Chen et~al.(2016)Chen, Sun, Tu, Lin, and Liu}]{rnn-2}
Huimin Chen, Maosong Sun, Cunchao Tu, Yankai Lin, and Zhiyuan Liu.
\newblock Neural sentiment classification with user and product attention.
\newblock In \emph{Proceedings of the 2016 Conference on Empirical Methods in
  Natural Language Processing}, pages 1650--1659, Austin, Texas. Association
  for Computational Linguistics.
\newblock

\bibitem[{Chen et~al.(2017)Chen, Sun, Bing, and Yang}]{rnn-4.4}
Peng Chen, Zhongqian Sun, Lidong Bing, and Wei Yang.
\newblock Recurrent attention network on memory for aspect sentiment analysis.
\newblock In \emph{Proceedings of the 2017 Conference on Empirical Methods in
  Natural Language Processing}, pages 452--461, Copenhagen, Denmark.
  Association for Computational Linguistics.
\newblock

\bibitem[{Chen and Qian(2019)}]{capsnet-1}
Zhuang Chen and Tieyun Qian.
\newblock Transfer capsule network for aspect level sentiment classification.
\newblock In \emph{Proceedings of the 57th Annual Meeting of the Association
  for Computational Linguistics}, pages 547--556, Florence, Italy. Association
  for Computational Linguistics.
\newblock

\bibitem[{Cliche(2017)}]{rnn-cnn-2}
Mathieu Cliche.
\newblock {BB}{\_}twtr at {S}em{E}val-2017 task 4: Twitter sentiment analysis
  with {CNN}s and {LSTM}s.
\newblock In \emph{Proceedings of the 11th International Workshop on Semantic
  Evaluation ({S}em{E}val-2017)}, pages 573--580, Vancouver, Canada.
  Association for Computational Linguistics.
\newblock

\bibitem[{Cortis et~al.(2017)Cortis, Freitas, Daudert, Huerlimann, Zarrouk,
  Handschuh, and Davis}]{absa-dataset-2}
Keith Cortis, Andr{\'e} Freitas, Tobias Daudert, Manuela Huerlimann, Manel
  Zarrouk, Siegfried Handschuh, and Brian Davis.
\newblock {S}em{E}val-2017 task 5: Fine-grained sentiment analysis on financial
  microblogs and news.
\newblock In \emph{Proceedings of the 11th International Workshop on Semantic
  Evaluation ({S}em{E}val-2017)}, pages 519--535, Vancouver, Canada.
  Association for Computational Linguistics.
\newblock

\bibitem[{Devlin et~al.(2019)Devlin, Chang, Lee, and Toutanova}]{bert-orig}
Jacob Devlin, Ming-Wei Chang, Kenton Lee, and Kristina Toutanova.
\newblock {BERT}: Pre-training of deep bidirectional transformers for language
  understanding.
\newblock In \emph{Proceedings of the 2019 Conference of the North {A}merican
  Chapter of the Association for Computational Linguistics: Human Language
  Technologies, Volume 1 (Long and Short Papers)}, pages 4171--4186,
  Minneapolis, Minnesota. Association for Computational Linguistics.
\newblock

\bibitem[{Dong et~al.(2014)Dong, Wei, Tan, Tang, Zhou, and Xu}]{absa-dataset-5}
Li~Dong, Furu Wei, Chuanqi Tan, Duyu Tang, Ming Zhou, and Ke~Xu.
\newblock Adaptive recursive neural network for target-dependent twitter
  sentiment classification.
\newblock In \emph{Proceedings of the 52nd Annual Meeting of the Association
  for Computational Linguistics (Volume 2: Short Papers)}, pages 49--54,
  Baltimore, Maryland. Association for Computational Linguistics.
\newblock

\bibitem[{Dong and de~Melo(2018)}]{cnn-5}
Xin Dong and Gerard de~Melo.
\newblock A helping hand: Transfer learning for deep sentiment analysis.
\newblock In \emph{Proceedings of the 56th Annual Meeting of the Association
  for Computational Linguistics (Volume 1: Long Papers)}, pages 2524--2534,
  Melbourne, Australia. Association for Computational Linguistics.
\newblock

\bibitem[{Ekman(1992)}]{er-types}
Paul Ekman.
\newblock An argument for basic emotions.
\newblock \emph{Cognition \& emotion}, 6(3-4):169--200.

\bibitem[{Felbo et~al.(2017)Felbo, Mislove, S{\o}gaard, Rahwan, and
  Lehmann}]{rnn-5.1}
Bjarke Felbo, Alan Mislove, Anders S{\o}gaard, Iyad Rahwan, and Sune Lehmann.
\newblock Using millions of emoji occurrences to learn any-domain
  representations for detecting sentiment, emotion and sarcasm.
\newblock In \emph{Proceedings of the 2017 Conference on Empirical Methods in
  Natural Language Processing}, pages 1615--1625, Copenhagen, Denmark.
  Association for Computational Linguistics.
\newblock

\bibitem[{Feng and Hirst(2013)}]{aa-dataset-novel9}
Vanessa~Wei Feng and Graeme Hirst.
\newblock {Patterns of local discourse coherence as a feature for authorship
  attribution}.
\newblock \emph{Literary and Linguistic Computing}, 29(2):191--198.
\newblock

\bibitem[{Ferracane et~al.(2017)Ferracane, Wang, and Mooney}]{cnn-4}
Elisa Ferracane, Su~Wang, and Raymond Mooney.
\newblock Leveraging discourse information effectively for authorship
  attribution.
\newblock In \emph{Proceedings of the Eighth International Joint Conference on
  Natural Language Processing (Volume 1: Long Papers)}, pages 584--593, Taipei,
  Taiwan. Asian Federation of Natural Language Processing.
\newblock

\bibitem[{Flekova et~al.(2016)Flekova, Preo{\c{t}}iuc-Pietro, and
  Ungar}]{pos-1}
Lucie Flekova, Daniel Preo{\c{t}}iuc-Pietro, and Lyle Ungar.
\newblock Exploring stylistic variation with age and income on twitter.
\newblock In \emph{Proceedings of the 54th Annual Meeting of the Association
  for Computational Linguistics (Volume 2: Short Papers)}, pages 313--319,
  Berlin, Germany. Association for Computational Linguistics.
\newblock

\bibitem[{Gao et~al.(2018)Gao, Choi, Choi, and Zettlemoyer}]{rnn-5.2}
Ge~Gao, Eunsol Choi, Yejin Choi, and Luke Zettlemoyer.
\newblock Neural metaphor detection in context.
\newblock In \emph{Proceedings of the 2018 Conference on Empirical Methods in
  Natural Language Processing}, pages 607--613, Brussels, Belgium. Association
  for Computational Linguistics.
\newblock

\bibitem[{Ghosal et~al.(2019)Ghosal, Majumder, Poria, Chhaya, and
  Gelbukh}]{gcn-2}
Deepanway Ghosal, Navonil Majumder, Soujanya Poria, Niyati Chhaya, and
  Alexander Gelbukh.
\newblock {D}ialogue{GCN}: A graph convolutional neural network for emotion
  recognition in conversation.
\newblock In \emph{Proceedings of the 2019 Conference on Empirical Methods in
  Natural Language Processing and the 9th International Joint Conference on
  Natural Language Processing (EMNLP-IJCNLP)}, pages 154--164, Hong Kong,
  China. Association for Computational Linguistics.
\newblock

\bibitem[{Goodfellow et~al.(2014)Goodfellow, Pouget{-}Abadie, Mirza, Xu,
  Warde{-}Farley, Ozair, Courville, and Bengio}]{adversarial-orig}
Ian~J. Goodfellow, Jean Pouget{-}Abadie, Mehdi Mirza, Bing Xu, David
  Warde{-}Farley, Sherjil Ozair, Aaron~C. Courville, and Yoshua Bengio.
\newblock Generative adversarial nets.
\newblock In \emph{Advances in Neural Information Processing Systems 27: Annual
  Conference on Neural Information Processing Systems 2014, December 8-13 2014,
  Montreal, Quebec, Canada}, pages 2672--2680.
\newblock

\bibitem[{Hasan and Ng(2013)}]{sd-orig-2}
Kazi~Saidul Hasan and Vincent Ng.
\newblock Stance classification of ideological debates: Data, models, features,
  and constraints.
\newblock In \emph{Proceedings of the Sixth International Joint Conference on
  Natural Language Processing}, pages 1348--1356, Nagoya, Japan. Asian
  Federation of Natural Language Processing.
\newblock

\bibitem[{Hazarika et~al.(2018{\natexlab{a}})Hazarika, Poria, Mihalcea,
  Cambria, and Zimmermann}]{memnn-rnn-cnn-1}
Devamanyu Hazarika, Soujanya Poria, Rada Mihalcea, Erik Cambria, and Roger
  Zimmermann.
\newblock {ICON}: Interactive conversational memory network for multimodal
  emotion detection.
\newblock In \emph{Proceedings of the 2018 Conference on Empirical Methods in
  Natural Language Processing}, pages 2594--2604, Brussels, Belgium.
  Association for Computational Linguistics.
\newblock

\bibitem[{Hazarika et~al.(2018{\natexlab{b}})Hazarika, Poria, Zadeh, Cambria,
  Morency, and Zimmermann}]{memnn-rnn-1}
Devamanyu Hazarika, Soujanya Poria, Amir Zadeh, Erik Cambria, Louis-Philippe
  Morency, and Roger Zimmermann.
\newblock Conversational memory network for emotion recognition in dyadic
  dialogue videos.
\newblock In \emph{Proceedings of the 2018 Conference of the North {A}merican
  Chapter of the Association for Computational Linguistics: Human Language
  Technologies, Volume 1 (Long Papers)}, pages 2122--2132, New Orleans,
  Louisiana. Association for Computational Linguistics.
\newblock

\bibitem[{Herrmann et~al.(2015)Herrmann, van Dalen-Oskam, and
  Sch{\"o}ch}]{style-hist-2}
J~Berenike Herrmann, Karina van Dalen-Oskam, and Christof Sch{\"o}ch.
\newblock Revisiting style, a key concept in literary studies.
\newblock \emph{Journal of literary theory}, 9(1):25--52.

\bibitem[{Hihi and Bengio(1995)}]{hiernn}
Salah~El Hihi and Yoshua Bengio.
\newblock Hierarchical recurrent neural networks for long-term dependencies.
\newblock In \emph{Advances in Neural Information Processing Systems 8, NIPS,
  Denver, CO, USA, November 27-30, 1995}, pages 493--499. {MIT} Press.
\newblock

\bibitem[{Hoover et~al.(2019)Hoover, Portillo-Wightman, Yeh, Havaldar, Davani,
  Lin, Kennedy, Atari, Kamel, Mendlen et~al.}]{sd-dataset-4}
Joseph Hoover, Gwenyth Portillo-Wightman, Leigh Yeh, Shreya Havaldar,
  Aida~Mostafazadeh Davani, Ying Lin, Brendan Kennedy, Mohammad Atari, Zahra
  Kamel, Madelyn Mendlen, et~al.
\newblock Moral foundations twitter corpus: A collection of 35k tweets
  annotated for moral sentiment.

\bibitem[{Hu(2019)}]{rnn-4.3}
Shengli Hu.
\newblock Detecting concealed information in text and speech.
\newblock In \emph{Proceedings of the 57th Annual Meeting of the Association
  for Computational Linguistics}, pages 402--412, Florence, Italy. Association
  for Computational Linguistics.
\newblock

\bibitem[{Huang and Carley(2019)}]{gat-1}
Binxuan Huang and Kathleen Carley.
\newblock Syntax-aware aspect level sentiment classification with graph
  attention networks.
\newblock In \emph{Proceedings of the 2019 Conference on Empirical Methods in
  Natural Language Processing and the 9th International Joint Conference on
  Natural Language Processing (EMNLP-IJCNLP)}, pages 5469--5477, Hong Kong,
  China. Association for Computational Linguistics.
\newblock

\bibitem[{Jafariakinabad et~al.(2020)Jafariakinabad, Tarnpradab, and
  Hua}]{aa-4}
Fereshteh Jafariakinabad, Sansiri Tarnpradab, and Kien~A. Hua.
\newblock Syntactic neural model for authorship attribution.
\newblock In \emph{Proceedings of the Thirty-Third International Florida
  Artificial Intelligence Research Society Conference, Originally to be held in
  North Miami Beach, Florida, USA, May 17-20, 2020}, pages 234--239. {AAAI}
  Press.
\newblock

\bibitem[{Jiang et~al.(2019)Jiang, Chen, Xu, Ao, and Yang}]{capsnet-2}
Qingnan Jiang, Lei Chen, Ruifeng Xu, Xiang Ao, and Min Yang.
\newblock A challenge dataset and effective models for aspect-based sentiment
  analysis.
\newblock In \emph{Proceedings of the 2019 Conference on Empirical Methods in
  Natural Language Processing and the 9th International Joint Conference on
  Natural Language Processing (EMNLP-IJCNLP)}, pages 6280--6285, Hong Kong,
  China. Association for Computational Linguistics.
\newblock

\bibitem[{Johansson(2019)}]{ap-1}
Fredrik Johansson.
\newblock Supervised classification of twitter accounts based on textual
  content of tweets.
\newblock In \emph{Working Notes of {CLEF} 2019 - Conference and Labs of the
  Evaluation Forum, Lugano, Switzerland, September 9-12, 2019}, volume 2380 of
  \emph{{CEUR} Workshop Proceedings}. CEUR-WS.org.
\newblock

\bibitem[{Joo and Hwang(2019)}]{ap-3}
Youngjun Joo and Inchon Hwang.
\newblock Author profiling on social media: An ensemble learning approach using
  various features.
\newblock In \emph{Working Notes of {CLEF} 2019 - Conference and Labs of the
  Evaluation Forum, Lugano, Switzerland, September 9-12, 2019}, volume 2380 of
  \emph{{CEUR} Workshop Proceedings}. CEUR-WS.org.
\newblock

\bibitem[{Juola(2006)}]{aa-orig}
Patrick Juola.
\newblock Authorship attribution.
\newblock \emph{Found. Trends Inf. Retr.}, 1(3):233--334.
\newblock

\bibitem[{Juola(2012)}]{pan-12-report}
Patrick Juola.
\newblock An overview of the traditional authorship attribution subtask.
\newblock In \emph{{CLEF} 2012 Evaluation Labs and Workshop, Online Working
  Notes, Rome, Italy, September 17-20, 2012}, volume 1178 of \emph{{CEUR}
  Workshop Proceedings}. CEUR-WS.org.
\newblock

\bibitem[{Kestemont et~al.(2019)Kestemont, Stamatatos, Manjavacas, Daelemans,
  Potthast, and Stein}]{pan-19-report}
Mike Kestemont, Efstathios Stamatatos, Enrique Manjavacas, Walter Daelemans,
  Martin Potthast, and Benno Stein.
\newblock Overview of the cross-domain authorship attribution task at {PAN}
  2019.
\newblock In \emph{Working Notes of {CLEF} 2019 - Conference and Labs of the
  Evaluation Forum, Lugano, Switzerland, September 9-12, 2019}, volume 2380 of
  \emph{{CEUR} Workshop Proceedings}. CEUR-WS.org.
\newblock

\bibitem[{Kestemont et~al.(2016)Kestemont, Stover, Koppel, Karsdorp, and
  Daelemans}]{style-hist-1}
Mike Kestemont, Justin~Anthony Stover, Moshe Koppel, Folgert Karsdorp, and
  Walter Daelemans.
\newblock Authenticating the writings of julius caesar.
\newblock \emph{Expert Syst. Appl.}, 63:86--96.
\newblock

\bibitem[{Khodak et~al.(2018)Khodak, Saunshi, and Vodrahalli}]{id-dataset-1}
Mikhail Khodak, Nikunj Saunshi, and Kiran Vodrahalli.
\newblock A large self-annotated corpus for sarcasm.
\newblock In \emph{Proceedings of the Eleventh International Conference on
  Language Resources and Evaluation ({LREC} 2018)}, Miyazaki, Japan. European
  Language Resources Association (ELRA).
\newblock

\bibitem[{Kipf and Welling(2017)}]{gcn-orig}
Thomas~N. Kipf and Max Welling.
\newblock Semi-supervised classification with graph convolutional networks.
\newblock In \emph{5th International Conference on Learning Representations,
  {ICLR} 2017, Toulon, France, April 24-26, 2017, Conference Track
  Proceedings}. OpenReview.net.
\newblock

\bibitem[{K{\"{u}}{\c{c}}{\"{u}}k and Can(2020)}]{sd-review}
Dilek K{\"{u}}{\c{c}}{\"{u}}k and Fazli Can.
\newblock Stance detection: {A} survey.
\newblock \emph{{ACM} Comput. Surv.}, 53(1):12:1--12:37.
\newblock

\bibitem[{Kumar et~al.(2017)Kumar, Sethi, Akhtar, Ekbal, Biemann, and
  Bhattacharyya}]{nlm-2}
Abhishek Kumar, Abhishek Sethi, Md~Shad Akhtar, Asif Ekbal, Chris Biemann, and
  Pushpak Bhattacharyya.
\newblock {IITPB} at {S}em{E}val-2017 task 5: Sentiment prediction in financial
  text.
\newblock In \emph{Proceedings of the 11th International Workshop on Semantic
  Evaluation ({S}em{E}val-2017)}, pages 894--898, Vancouver, Canada.
  Association for Computational Linguistics.
\newblock

\bibitem[{LeCun et~al.(2015)LeCun, Bengio, and Hinton}]{dl-lecun}
Yann LeCun, Yoshua Bengio, and Geoffrey Hinton.
\newblock Deep learning.
\newblock \emph{nature}, 521(7553):436--444.

\bibitem[{LeCun et~al.(1999)LeCun, Haffner, Bottou, and Bengio}]{cnn-orig}
Yann LeCun, Patrick Haffner, L{\'{e}}on Bottou, and Yoshua Bengio.
\newblock Object recognition with gradient-based learning.
\newblock In \emph{Shape, Contour and Grouping in Computer Vision}, volume 1681
  of \emph{Lecture Notes in Computer Science}, page 319. Springer.
\newblock

\bibitem[{Li et~al.(2018)Li, Bing, Lam, and Shi}]{rnn-cnn-3}
Xin Li, Lidong Bing, Wai Lam, and Bei Shi.
\newblock Transformation networks for target-oriented sentiment classification.
\newblock In \emph{Proceedings of the 56th Annual Meeting of the Association
  for Computational Linguistics (Volume 1: Long Papers)}, pages 946--956,
  Melbourne, Australia. Association for Computational Linguistics.
\newblock

\bibitem[{Mao et~al.(2018)Mao, Lin, and Guerin}]{lexicon-4}
Rui Mao, Chenghua Lin, and Frank Guerin.
\newblock Word embedding and {W}ord{N}et based metaphor identification and
  interpretation.
\newblock In \emph{Proceedings of the 56th Annual Meeting of the Association
  for Computational Linguistics (Volume 1: Long Papers)}, pages 1222--1231,
  Melbourne, Australia. Association for Computational Linguistics.
\newblock

\bibitem[{Markov et~al.(2017)Markov, Stamatatos, and Sidorov}]{ngram-5}
Ilia Markov, Efstathios Stamatatos, and Grigori Sidorov.
\newblock Improving cross-topic authorship attribution: The role of
  pre-processing.
\newblock In \emph{International Conference on Computational Linguistics and
  Intelligent Text Processing}, pages 289--302. Springer.

\bibitem[{Martinc et~al.(2019)Martinc, Skrlj, and Pollak}]{cp-1}
Matej Martinc, Blaz Skrlj, and Senja Pollak.
\newblock Who is hot and who is not? profiling celebs on twitter.
\newblock In \emph{Working Notes of {CLEF} 2019 - Conference and Labs of the
  Evaluation Forum, Lugano, Switzerland, September 9-12, 2019}, volume 2380 of
  \emph{{CEUR} Workshop Proceedings}. CEUR-WS.org.
\newblock

\bibitem[{Mihalcea and Strapparava(2005)}]{task-specific-1}
Rada Mihalcea and Carlo Strapparava.
\newblock Making computers laugh: Investigations in automatic humor
  recognition.
\newblock In \emph{Proceedings of Human Language Technology Conference and
  Conference on Empirical Methods in Natural Language Processing}, pages
  531--538, Vancouver, British Columbia, Canada. Association for Computational
  Linguistics.
\newblock

\bibitem[{Mikolov et~al.(2013)Mikolov, Chen, Corrado, and Dean}]{skipgram-orig}
Tomas Mikolov, Kai Chen, Greg Corrado, and Jeffrey Dean.
\newblock Efficient estimation of word representations in vector space.
\newblock In \emph{1st International Conference on Learning Representations,
  {ICLR} 2013, Scottsdale, Arizona, USA, May 2-4, 2013, Workshop Track
  Proceedings}.
\newblock

\bibitem[{Miller(1998)}]{wordnet}
George~A Miller.
\newblock \emph{WordNet: An electronic lexical database}.
\newblock MIT press.

\bibitem[{Mohammad(2018)}]{nrc-orig}
Saif Mohammad.
\newblock Obtaining reliable human ratings of valence, arousal, and dominance
  for 20,000 {E}nglish words.
\newblock In \emph{Proceedings of the 56th Annual Meeting of the Association
  for Computational Linguistics (Volume 1: Long Papers)}, pages 174--184,
  Melbourne, Australia. Association for Computational Linguistics.
\newblock

\bibitem[{Mohammad et~al.(2016{\natexlab{a}})Mohammad, Kiritchenko, Sobhani,
  Zhu, and Cherry}]{sd-dataset-3}
Saif Mohammad, Svetlana Kiritchenko, Parinaz Sobhani, Xiaodan Zhu, and Colin
  Cherry.
\newblock A dataset for detecting stance in tweets.
\newblock In \emph{Proceedings of the Tenth International Conference on
  Language Resources and Evaluation ({LREC}'16)}, pages 3945--3952,
  Portoro{\v{z}}, Slovenia. European Language Resources Association (ELRA).
\newblock

\bibitem[{Mohammad et~al.(2016{\natexlab{b}})Mohammad, Kiritchenko, Sobhani,
  Zhu, and Cherry}]{sd-dataset-1}
Saif Mohammad, Svetlana Kiritchenko, Parinaz Sobhani, Xiaodan Zhu, and Colin
  Cherry.
\newblock {S}em{E}val-2016 task 6: Detecting stance in tweets.
\newblock In \emph{Proceedings of the 10th International Workshop on Semantic
  Evaluation ({S}em{E}val-2016)}, pages 31--41, San Diego, California.
  Association for Computational Linguistics.
\newblock

\bibitem[{Mohammad et~al.(2016{\natexlab{c}})Mohammad, Shutova, and
  Turney}]{md-dataset-moh}
Saif Mohammad, Ekaterina Shutova, and Peter Turney.
\newblock Metaphor as a medium for emotion: An empirical study.
\newblock In \emph{Proceedings of the Fifth Joint Conference on Lexical and
  Computational Semantics}, pages 23--33, Berlin, Germany. Association for
  Computational Linguistics.
\newblock

\bibitem[{Mohammad and Turney(2010)}]{lexicon-2.1}
Saif Mohammad and Peter Turney.
\newblock Emotions evoked by common words and phrases: Using mechanical turk to
  create an emotion lexicon.
\newblock In \emph{Proceedings of the {NAACL} {HLT} 2010 Workshop on
  Computational Approaches to Analysis and Generation of Emotion in Text},
  pages 26--34, Los Angeles, CA. Association for Computational Linguistics.
\newblock

\bibitem[{Mohammad and Turney(2013)}]{lexicon-2.2}
Saif~M Mohammad and Peter~D Turney.
\newblock Crowdsourcing a word--emotion association lexicon.
\newblock \emph{Computational Intelligence}, 29(3):436--465.

\bibitem[{Mohler et~al.(2016)Mohler, Brunson, Rink, and
  Tomlinson}]{md-dataset-lcc}
Michael Mohler, Mary Brunson, Bryan Rink, and Marc Tomlinson.
\newblock Introducing the {LCC} metaphor datasets.
\newblock In \emph{Proceedings of the Tenth International Conference on
  Language Resources and Evaluation ({LREC}'16)}, pages 4221--4227,
  Portoro{\v{z}}, Slovenia. European Language Resources Association (ELRA).
\newblock

\bibitem[{Muttenthaler et~al.(2019)Muttenthaler, Lucas, and Amann}]{ngram-2}
Lukas Muttenthaler, Gordon Lucas, and Janek Amann.
\newblock Authorship attribution in fan-fictional texts given variable length
  character and word n-grams.
\newblock In \emph{Working Notes of {CLEF} 2019 - Conference and Labs of the
  Evaluation Forum, Lugano, Switzerland, September 9-12, 2019}, volume 2380 of
  \emph{{CEUR} Workshop Proceedings}. CEUR-WS.org.
\newblock

\bibitem[{Niu et~al.(2017)Niu, Martindale, and Carpuat}]{pca-1}
Xing Niu, Marianna Martindale, and Marine Carpuat.
\newblock A study of style in machine translation: Controlling the formality of
  machine translation output.
\newblock In \emph{Proceedings of the 2017 Conference on Empirical Methods in
  Natural Language Processing}, pages 2814--2819, Copenhagen, Denmark.
  Association for Computational Linguistics.
\newblock

\bibitem[{Ousidhoum et~al.(2019)Ousidhoum, Lin, Zhang, Song, and
  Yeung}]{od-dataset-2}
Nedjma Ousidhoum, Zizheng Lin, Hongming Zhang, Yangqiu Song, and Dit-Yan Yeung.
\newblock Multilingual and multi-aspect hate speech analysis.
\newblock In \emph{Proceedings of the 2019 Conference on Empirical Methods in
  Natural Language Processing and the 9th International Joint Conference on
  Natural Language Processing (EMNLP-IJCNLP)}, pages 4667--4676, Hong Kong,
  China. Association for Computational Linguistics.
\newblock

\bibitem[{Pang and Lee(2005)}]{rt-orig}
Bo~Pang and Lillian Lee.
\newblock Seeing stars: Exploiting class relationships for sentiment
  categorization with respect to rating scales.
\newblock In \emph{Proceedings of the 43rd Annual Meeting of the Association
  for Computational Linguistics ({ACL}{'}05)}, pages 115--124, Ann Arbor,
  Michigan. Association for Computational Linguistics.
\newblock

\bibitem[{Pang and Lee(2008)}]{absa-orig}
Bo~Pang and Lillian Lee.
\newblock Opinion mining and sentiment analysis.
\newblock \emph{Found. Trends Inf. Retr.}, 2(1-2):1--135.
\newblock

\bibitem[{Pang et~al.(2002)Pang, Lee, and Vaithyanathan}]{sa-orig}
Bo~Pang, Lillian Lee, and Shivakumar Vaithyanathan.
\newblock Thumbs up? sentiment classification using machine learning
  techniques.
\newblock In \emph{Proceedings of the 2002 Conference on Empirical Methods in
  Natural Language Processing ({EMNLP} 2002)}, pages 79--86. Association for
  Computational Linguistics.
\newblock

\bibitem[{Pardo and Rosso(2019)}]{pan-19-ap-orig}
Francisco M.~Rangel Pardo and Paolo Rosso.
\newblock Overview of the 7th author profiling task at {PAN} 2019: Bots and
  gender profiling in twitter.
\newblock In \emph{Working Notes of {CLEF} 2019 - Conference and Labs of the
  Evaluation Forum, Lugano, Switzerland, September 9-12, 2019}, volume 2380 of
  \emph{{CEUR} Workshop Proceedings}. CEUR-WS.org.
\newblock

\bibitem[{P{\'e}rez-Rosas et~al.(2013)P{\'e}rez-Rosas, Mihalcea, and
  Morency}]{er-dataset-moud}
Ver{\'o}nica P{\'e}rez-Rosas, Rada Mihalcea, and Louis-Philippe Morency.
\newblock Utterance-level multimodal sentiment analysis.
\newblock In \emph{Proceedings of the 51st Annual Meeting of the Association
  for Computational Linguistics (Volume 1: Long Papers)}, pages 973--982,
  Sofia, Bulgaria. Association for Computational Linguistics.
\newblock

\bibitem[{Petr{\'{\i}}k and Chud{\'{a}}(2019)}]{ap-6}
Juraj Petr{\'{\i}}k and Daniela Chud{\'{a}}.
\newblock Bots and gender profiling with convolutional hierarchical recurrent
  neural network.
\newblock In \emph{Working Notes of {CLEF} 2019 - Conference and Labs of the
  Evaluation Forum, Lugano, Switzerland, September 9-12, 2019}, volume 2380 of
  \emph{{CEUR} Workshop Proceedings}. CEUR-WS.org.
\newblock

\bibitem[{Polignano et~al.(2019)Polignano, de~Pinto, Lops, and Semeraro}]{ap-5}
Marco Polignano, Marco~Giuseppe de~Pinto, Pasquale Lops, and Giovanni Semeraro.
\newblock Identification of bot accounts in twitter using 2d cnns on
  user-generated contents.
\newblock In \emph{Working Notes of {CLEF} 2019 - Conference and Labs of the
  Evaluation Forum, Lugano, Switzerland, September 9-12, 2019}, volume 2380 of
  \emph{{CEUR} Workshop Proceedings}. CEUR-WS.org.
\newblock

\bibitem[{Pontiki et~al.(2016)Pontiki, Galanis, Papageorgiou, Androutsopoulos,
  Manandhar, AL-Smadi, Al-Ayyoub, Zhao, Qin, De~Clercq, Hoste, Apidianaki,
  Tannier, Loukachevitch, Kotelnikov, Bel, Jim{\'e}nez-Zafra, and
  Eryi{\u{g}}it}]{absa-dataset-4}
Maria Pontiki, Dimitris Galanis, Haris Papageorgiou, Ion Androutsopoulos,
  Suresh Manandhar, Mohammad AL-Smadi, Mahmoud Al-Ayyoub, Yanyan Zhao, Bing
  Qin, Orph{\'e}e De~Clercq, V{\'e}ronique Hoste, Marianna Apidianaki, Xavier
  Tannier, Natalia Loukachevitch, Evgeniy Kotelnikov, Nuria Bel,
  Salud~Mar{\'\i}a Jim{\'e}nez-Zafra, and G{\"u}l{\c{s}}en Eryi{\u{g}}it.
\newblock {S}em{E}val-2016 task 5: Aspect based sentiment analysis.
\newblock In \emph{Proceedings of the 10th International Workshop on Semantic
  Evaluation ({S}em{E}val-2016)}, pages 19--30, San Diego, California.
  Association for Computational Linguistics.
\newblock

\bibitem[{Pontiki et~al.(2015)Pontiki, Galanis, Papageorgiou, Manandhar, and
  Androutsopoulos}]{absa-dataset-3}
Maria Pontiki, Dimitris Galanis, Haris Papageorgiou, Suresh Manandhar, and Ion
  Androutsopoulos.
\newblock {S}em{E}val-2015 task 12: Aspect based sentiment analysis.
\newblock In \emph{Proceedings of the 9th International Workshop on Semantic
  Evaluation ({S}em{E}val 2015)}, pages 486--495, Denver, Colorado. Association
  for Computational Linguistics.
\newblock

\bibitem[{Pontiki et~al.(2014)Pontiki, Galanis, Pavlopoulos, Papageorgiou,
  Androutsopoulos, and Manandhar}]{absa-dataset-1}
Maria Pontiki, Dimitris Galanis, John Pavlopoulos, Harris Papageorgiou, Ion
  Androutsopoulos, and Suresh Manandhar.
\newblock {S}em{E}val-2014 task 4: Aspect based sentiment analysis.
\newblock In \emph{Proceedings of the 8th International Workshop on Semantic
  Evaluation ({S}em{E}val 2014)}, pages 27--35, Dublin, Ireland. Association
  for Computational Linguistics.
\newblock

\bibitem[{Poria et~al.(2019)Poria, Hazarika, Majumder, Naik, Cambria, and
  Mihalcea}]{er-dataset-meld}
Soujanya Poria, Devamanyu Hazarika, Navonil Majumder, Gautam Naik, Erik
  Cambria, and Rada Mihalcea.
\newblock {MELD}: A multimodal multi-party dataset for emotion recognition in
  conversations.
\newblock In \emph{Proceedings of the 57th Annual Meeting of the Association
  for Computational Linguistics}, pages 527--536, Florence, Italy. Association
  for Computational Linguistics.
\newblock

\bibitem[{Preo{\c{t}}iuc-Pietro and Devlin~Marier(2019)}]{nlm-4}
Daniel Preo{\c{t}}iuc-Pietro and Rita Devlin~Marier.
\newblock Analyzing linguistic differences between owner and staff attributed
  tweets.
\newblock In \emph{Proceedings of the 57th Annual Meeting of the Association
  for Computational Linguistics}, pages 2848--2853, Florence, Italy.
  Association for Computational Linguistics.
\newblock

\bibitem[{Preo{\c{t}}iuc-Pietro et~al.(2019)Preo{\c{t}}iuc-Pietro, Gaman, and
  Aletras}]{pos-2}
Daniel Preo{\c{t}}iuc-Pietro, Mihaela Gaman, and Nikolaos Aletras.
\newblock Automatically identifying complaints in social media.
\newblock In \emph{Proceedings of the 57th Annual Meeting of the Association
  for Computational Linguistics}, pages 5008--5019, Florence, Italy.
  Association for Computational Linguistics.
\newblock

\bibitem[{Qian et~al.(2017)Qian, Huang, Lei, and Zhu}]{rnn-4.1}
Qiao Qian, Minlie Huang, Jinhao Lei, and Xiaoyan Zhu.
\newblock Linguistically regularized {LSTM} for sentiment classification.
\newblock In \emph{Proceedings of the 55th Annual Meeting of the Association
  for Computational Linguistics (Volume 1: Long Papers)}, pages 1679--1689,
  Vancouver, Canada. Association for Computational Linguistics.
\newblock

\bibitem[{Rao and Tetreault(2018)}]{fc-dataset-1}
Sudha Rao and Joel Tetreault.
\newblock Dear sir or madam, may {I} introduce the {GYAFC} dataset: Corpus,
  benchmarks and metrics for formality style transfer.
\newblock In \emph{Proceedings of the 2018 Conference of the North {A}merican
  Chapter of the Association for Computational Linguistics: Human Language
  Technologies, Volume 1 (Long Papers)}, pages 129--140, New Orleans,
  Louisiana. Association for Computational Linguistics.
\newblock

\bibitem[{Rei et~al.(2017)Rei, Bulat, Kiela, and Shutova}]{nlm-3}
Marek Rei, Luana Bulat, Douwe Kiela, and Ekaterina Shutova.
\newblock Grasping the finer point: A supervised similarity network for
  metaphor detection.
\newblock In \emph{Proceedings of the 2017 Conference on Empirical Methods in
  Natural Language Processing}, pages 1537--1546, Copenhagen, Denmark.
  Association for Computational Linguistics.
\newblock

\bibitem[{Rouvier(2017)}]{rnn-cnn-1}
Mickael Rouvier.
\newblock {LIA} at {S}em{E}val-2017 task 4: An ensemble of neural networks for
  sentiment classification.
\newblock In \emph{Proceedings of the 11th International Workshop on Semantic
  Evaluation ({S}em{E}val-2017)}, pages 760--765, Vancouver, Canada.
  Association for Computational Linguistics.
\newblock

\bibitem[{Ruder et~al.(2016)Ruder, Ghaffari, and Breslin}]{cnn-3}
Sebastian Ruder, Parsa Ghaffari, and John~G Breslin.
\newblock Character-level and multi-channel convolutional neural networks for
  large-scale authorship attribution.
\newblock \emph{arXiv preprint arXiv:1609.06686}.

\bibitem[{Sabour et~al.(2017)Sabour, Frosst, and Hinton}]{capsnet-orig}
Sara Sabour, Nicholas Frosst, and Geoffrey~E. Hinton.
\newblock Dynamic routing between capsules.
\newblock In \emph{Advances in Neural Information Processing Systems 30: Annual
  Conference on Neural Information Processing Systems 2017, 4-9 December 2017,
  Long Beach, CA, {USA}}, pages 3856--3866.
\newblock

\bibitem[{Saeidi et~al.(2016)Saeidi, Bouchard, Liakata, and
  Riedel}]{absa-review}
Marzieh Saeidi, Guillaume Bouchard, Maria Liakata, and Sebastian Riedel.
\newblock {S}enti{H}ood: Targeted aspect based sentiment analysis dataset for
  urban neighbourhoods.
\newblock In \emph{Proceedings of {COLING} 2016, the 26th International
  Conference on Computational Linguistics: Technical Papers}, pages 1546--1556,
  Osaka, Japan. The COLING 2016 Organizing Committee.
\newblock

\bibitem[{Sapkota et~al.(2015)Sapkota, Bethard, Montes, and Solorio}]{ngram-1}
Upendra Sapkota, Steven Bethard, Manuel Montes, and Thamar Solorio.
\newblock Not all character n-grams are created equal: A study in authorship
  attribution.
\newblock In \emph{Proceedings of the 2015 Conference of the North {A}merican
  Chapter of the Association for Computational Linguistics: Human Language
  Technologies}, pages 93--102, Denver, Colorado. Association for Computational
  Linguistics.
\newblock

\bibitem[{Sari et~al.(2017)Sari, Vlachos, and Stevenson}]{aa-2}
Yunita Sari, Andreas Vlachos, and Mark Stevenson.
\newblock Continuous n-gram representations for authorship attribution.
\newblock In \emph{Proceedings of the 15th Conference of the {E}uropean Chapter
  of the Association for Computational Linguistics: Volume 2, Short Papers},
  pages 267--273, Valencia, Spain. Association for Computational Linguistics.
\newblock

\bibitem[{Schler et~al.(2006)Schler, Koppel, Argamon, and
  Pennebaker}]{aa-dataset-blogs10}
Jonathan Schler, Moshe Koppel, Shlomo Argamon, and James~W. Pennebaker.
\newblock Effects of age and gender on blogging.
\newblock In \emph{Computational Approaches to Analyzing Weblogs, Papers from
  the 2006 {AAAI} Spring Symposium, Technical Report SS-06-03, Stanford,
  California, USA, March 27-29, 2006}, pages 199--205.
\newblock

\bibitem[{Schuster and Paliwal(1997)}]{birnn}
Mike Schuster and Kuldip~K Paliwal.
\newblock Bidirectional recurrent neural networks.
\newblock \emph{IEEE transactions on Signal Processing}, 45(11):2673--2681.

\bibitem[{Seroussi et~al.(2010)Seroussi, Zukerman, and
  Bohnert}]{aa-dataset-imdb62}
Yanir Seroussi, Ingrid Zukerman, and Fabian Bohnert.
\newblock Collaborative inference of sentiments from texts.
\newblock In \emph{Proceedings of the 18th International Conference on User
  Modeling, Adaptation, and Personalization}, UMAP'10, pages 195--206, Berlin,
  Heidelberg. Springer-Verlag.
\newblock

\bibitem[{Siddiqua et~al.(2019)Siddiqua, Chy, and Aono}]{sd-dataset-5}
Umme~Aymun Siddiqua, Abu~Nowshed Chy, and Masaki Aono.
\newblock {T}weet stance detection using an attention based neural ensemble
  model.
\newblock In \emph{Proceedings of the 2019 Conference of the North {A}merican
  Chapter of the Association for Computational Linguistics: Human Language
  Technologies, Volume 1 (Long and Short Papers)}, pages 1868--1873,
  Minneapolis, Minnesota. Association for Computational Linguistics.
\newblock

\bibitem[{Simaki et~al.(2017)Simaki, Skeppstedt, Paradis, Kerren, and
  Sahlgren}]{sd-dataset-2}
Vasiliki Simaki, Maria Skeppstedt, Carita Paradis, Andreas Kerren, and Magnus
  Sahlgren.
\newblock Annotating speaker stance in discourse : The brexit blog corpus.
\newblock

\bibitem[{Socher et~al.(2013)Socher, Perelygin, Wu, Chuang, Manning, Ng, and
  Potts}]{sst-orig}
Richard Socher, Alex Perelygin, Jean Wu, Jason Chuang, Christopher~D. Manning,
  Andrew Ng, and Christopher Potts.
\newblock Recursive deep models for semantic compositionality over a sentiment
  treebank.
\newblock In \emph{Proceedings of the 2013 Conference on Empirical Methods in
  Natural Language Processing}, pages 1631--1642, Seattle, Washington, USA.
  Association for Computational Linguistics.
\newblock

\bibitem[{Speer et~al.(2017)Speer, Chin, and Havasi}]{conceptnet-orig}
Robyn Speer, Joshua Chin, and Catherine Havasi.
\newblock Conceptnet 5.5: An open multilingual graph of general knowledge.
\newblock In \emph{Proceedings of the Thirty-First {AAAI} Conference on
  Artificial Intelligence, February 4-9, 2017, San Francisco, California,
  {USA}}, pages 4444--4451. {AAAI} Press.
\newblock

\bibitem[{Stamatatos(2008)}]{aa-dataset-ccat10}
Efstathios Stamatatos.
\newblock Author identification: Using text sampling to handle the class
  imbalance problem.
\newblock \emph{Inf. Process. Manage.}, 44(2):790--799.
\newblock

\bibitem[{Stamatatos(2009)}]{survey-past-1}
Efstathios Stamatatos.
\newblock A survey of modern authorship attribution methods.
\newblock \emph{J. Assoc. Inf. Sci. Technol.}, 60(3):538--556.
\newblock

\bibitem[{Steen et~al.(2010)Steen, Dorst, Herrmann, Kaal, and
  Krennmayr}]{md-dataset-vua}
Gerard~J Steen, Aletta~G Dorst, J~Berenike Herrmann, Anna~A Kaal, and Tina
  Krennmayr.
\newblock Metaphor in usage.
\newblock \emph{Cognitive Linguistics}, 21(4):765--796.

\bibitem[{Sun et~al.(2019)Sun, Zhang, Mensah, Mao, and Liu}]{gcn-1.1}
Kai Sun, Richong Zhang, Samuel Mensah, Yongyi Mao, and Xudong Liu.
\newblock Aspect-level sentiment analysis via convolution over dependency tree.
\newblock In \emph{Proceedings of the 2019 Conference on Empirical Methods in
  Natural Language Processing and the 9th International Joint Conference on
  Natural Language Processing (EMNLP-IJCNLP)}, pages 5679--5688, Hong Kong,
  China. Association for Computational Linguistics.
\newblock

\bibitem[{Tang et~al.(2016)Tang, Qin, and Liu}]{memnn-1}
Duyu Tang, Bing Qin, and Ting Liu.
\newblock Aspect level sentiment classification with deep memory network.
\newblock In \emph{Proceedings of the 2016 Conference on Empirical Methods in
  Natural Language Processing}, pages 214--224, Austin, Texas. Association for
  Computational Linguistics.
\newblock

\bibitem[{Tang et~al.(2019)Tang, Lu, Su, Ge, Song, Sun, and Luo}]{absa-2}
Jialong Tang, Ziyao Lu, Jinsong Su, Yubin Ge, Linfeng Song, Le~Sun, and Jiebo
  Luo.
\newblock Progressive self-supervised attention learning for aspect-level
  sentiment analysis.
\newblock In \emph{Proceedings of the 57th Annual Meeting of the Association
  for Computational Linguistics}, pages 557--566, Florence, Italy. Association
  for Computational Linguistics.
\newblock

\bibitem[{Thomas et~al.(2006)Thomas, Pang, and Lee}]{sd-orig-1}
Matt Thomas, Bo~Pang, and Lillian Lee.
\newblock Get out the vote: Determining support or opposition from
  congressional floor-debate transcripts.
\newblock In \emph{Proceedings of the 2006 Conference on Empirical Methods in
  Natural Language Processing}, pages 327--335, Sydney, Australia. Association
  for Computational Linguistics.
\newblock

\bibitem[{Tsvetkov et~al.(2014)Tsvetkov, Boytsov, Gershman, Nyberg, and
  Dyer}]{md-dataset-tsv}
Yulia Tsvetkov, Leonid Boytsov, Anatole Gershman, Eric Nyberg, and Chris Dyer.
\newblock Metaphor detection with cross-lingual model transfer.
\newblock In \emph{Proceedings of the 52nd Annual Meeting of the Association
  for Computational Linguistics (Volume 1: Long Papers)}, pages 248--258,
  Baltimore, Maryland. Association for Computational Linguistics.
\newblock

\bibitem[{Valencia et~al.(2019)Valencia, G{\'{o}}mez{-}Adorno, Rhodes, and
  Pineda}]{ap-2}
Alex I.~Valencia Valencia, Helena G{\'{o}}mez{-}Adorno, Christopher~Stephens
  Rhodes, and Gibran~Fuentes Pineda.
\newblock Bots and gender identification based on stylometry of tweet minimal
  structure and n-grams model.
\newblock In \emph{Working Notes of {CLEF} 2019 - Conference and Labs of the
  Evaluation Forum, Lugano, Switzerland, September 9-12, 2019}, volume 2380 of
  \emph{{CEUR} Workshop Proceedings}. CEUR-WS.org.
\newblock

\bibitem[{Vaswani et~al.(2017)Vaswani, Shazeer, Parmar, Uszkoreit, Jones,
  Gomez, Kaiser, and Polosukhin}]{transformer-orig}
Ashish Vaswani, Noam Shazeer, Niki Parmar, Jakob Uszkoreit, Llion Jones,
  Aidan~N. Gomez, Lukasz Kaiser, and Illia Polosukhin.
\newblock Attention is all you need.
\newblock In \emph{Advances in Neural Information Processing Systems 30: Annual
  Conference on Neural Information Processing Systems 2017, 4-9 December 2017,
  Long Beach, CA, {USA}}, pages 5998--6008.
\newblock

\bibitem[{Velickovic et~al.(2018)Velickovic, Cucurull, Casanova, Romero,
  Li{\`{o}}, and Bengio}]{gat-orig}
Petar Velickovic, Guillem Cucurull, Arantxa Casanova, Adriana Romero, Pietro
  Li{\`{o}}, and Yoshua Bengio.
\newblock Graph attention networks.
\newblock In \emph{6th International Conference on Learning Representations,
  {ICLR} 2018, Vancouver, BC, Canada, April 30 - May 3, 2018, Conference Track
  Proceedings}. OpenReview.net.
\newblock

\bibitem[{Volkova and Bachrach(2016)}]{lexicon-3}
Svitlana Volkova and Yoram Bachrach.
\newblock Inferring perceived demographics from user emotional tone and
  user-environment emotional contrast.
\newblock In \emph{Proceedings of the 54th Annual Meeting of the Association
  for Computational Linguistics (Volume 1: Long Papers)}, pages 1567--1578,
  Berlin, Germany. Association for Computational Linguistics.
\newblock

\bibitem[{Vu et~al.(2018)Vu, Nguyen, Vu, Nguyen, Catt, and Trenell}]{ngram-3}
Thanh Vu, Dat~Quoc Nguyen, Xuan-Son Vu, Dai~Quoc Nguyen, Michael Catt, and
  Michael Trenell.
\newblock {NIHRIO} at {S}em{E}val-2018 task 3: A simple and accurate neural
  network model for irony detection in twitter.
\newblock In \emph{Proceedings of The 12th International Workshop on Semantic
  Evaluation}, pages 525--530, New Orleans, Louisiana. Association for
  Computational Linguistics.
\newblock

\bibitem[{Weller and Seppi(2019)}]{transformer-1}
Orion Weller and Kevin Seppi.
\newblock Humor detection: A transformer gets the last laugh.
\newblock In \emph{Proceedings of the 2019 Conference on Empirical Methods in
  Natural Language Processing and the 9th International Joint Conference on
  Natural Language Processing (EMNLP-IJCNLP)}, pages 3621--3625, Hong Kong,
  China. Association for Computational Linguistics.
\newblock

\bibitem[{Weston et~al.(2015)Weston, Chopra, and Bordes}]{memnn-orig}
Jason Weston, Sumit Chopra, and Antoine Bordes.
\newblock Memory networks.
\newblock In \emph{3rd International Conference on Learning Representations,
  {ICLR} 2015, San Diego, CA, USA, May 7-9, 2015, Conference Track
  Proceedings}.
\newblock

\bibitem[{Wiebe et~al.(2005)Wiebe, Wilson, and Cardie}]{lexicon-1}
Janyce Wiebe, Theresa Wilson, and Claire Cardie.
\newblock Annotating expressions of opinions and emotions in language.
\newblock \emph{Language resources and evaluation}, 39(2-3):165--210.

\bibitem[{Wiegmann et~al.(2019)Wiegmann, Stein, and Potthast}]{cp-orig}
Matti Wiegmann, Benno Stein, and Martin Potthast.
\newblock Celebrity profiling.
\newblock In \emph{Proceedings of the 57th Annual Meeting of the Association
  for Computational Linguistics}, pages 2611--2618, Florence, Italy.
  Association for Computational Linguistics.
\newblock

\bibitem[{W{\"o}llmer et~al.(2013)W{\"o}llmer, Weninger, Knaup, Schuller, Sun,
  Sagae, and Morency}]{er-dataset-ictmmo}
Martin W{\"o}llmer, Felix Weninger, Tobias Knaup, Bj{\"o}rn Schuller, Congkai
  Sun, Kenji Sagae, and Louis-Philippe Morency.
\newblock Youtube movie reviews: Sentiment analysis in an audio-visual context.
\newblock \emph{IEEE Intelligent Systems}, 28(3):46--53.

\bibitem[{Xia et~al.(2018)Xia, Zhang, Yan, Chang, and Yu}]{capsnet-3}
Congying Xia, Chenwei Zhang, Xiaohui Yan, Yi~Chang, and Philip Yu.
\newblock Zero-shot user intent detection via capsule neural networks.
\newblock In \emph{Proceedings of the 2018 Conference on Empirical Methods in
  Natural Language Processing}, pages 3090--3099, Brussels, Belgium.
  Association for Computational Linguistics.
\newblock

\bibitem[{Xu et~al.(2018)Xu, Paris, Nepal, and Sparks}]{rnn-4.2}
Chang Xu, C{\'e}cile Paris, Surya Nepal, and Ross Sparks.
\newblock Cross-target stance classification with self-attention networks.
\newblock In \emph{Proceedings of the 56th Annual Meeting of the Association
  for Computational Linguistics (Volume 2: Short Papers)}, pages 778--783,
  Melbourne, Australia. Association for Computational Linguistics.
\newblock

\bibitem[{Xu et~al.(2019)Xu, Zhao, Yan, Zeng, Liang, and Sun}]{sa-1}
Jingjing Xu, Liang Zhao, Hanqi Yan, Qi~Zeng, Yun Liang, and Xu~Sun.
\newblock {L}exical{AT}: Lexical-based adversarial reinforcement training for
  robust sentiment classification.
\newblock In \emph{Proceedings of the 2019 Conference on Empirical Methods in
  Natural Language Processing and the 9th International Joint Conference on
  Natural Language Processing (EMNLP-IJCNLP)}, pages 5518--5527, Hong Kong,
  China. Association for Computational Linguistics.
\newblock

\bibitem[{Xue and Li(2018)}]{cnn-1}
Wei Xue and Tao Li.
\newblock Aspect based sentiment analysis with gated convolutional networks.
\newblock In \emph{Proceedings of the 56th Annual Meeting of the Association
  for Computational Linguistics (Volume 1: Long Papers)}, pages 2514--2523,
  Melbourne, Australia. Association for Computational Linguistics.
\newblock

\bibitem[{Yang et~al.(2017)Yang, Tu, Wang, Xu, and Chen}]{rnn-5.3}
Min Yang, Wenting Tu, Jingxuan Wang, Fei Xu, and Xiaojun Chen.
\newblock Attention based lstm for target dependent sentiment classification.
\newblock

\bibitem[{Zadeh et~al.(2016)Zadeh, Zellers, Pincus, and
  Morency}]{er-dataset-mosi}
Amir Zadeh, Rowan Zellers, Eli Pincus, and Louis-Philippe Morency.
\newblock Multimodal sentiment intensity analysis in videos: Facial gestures
  and verbal messages.
\newblock \emph{IEEE Intelligent Systems}, 31(6):82--88.

\bibitem[{Zampieri et~al.(2019)Zampieri, Malmasi, Nakov, Rosenthal, Farra, and
  Kumar}]{od-dataset-1}
Marcos Zampieri, Shervin Malmasi, Preslav Nakov, Sara Rosenthal, Noura Farra,
  and Ritesh Kumar.
\newblock Predicting the type and target of offensive posts in social media.
\newblock In \emph{Proceedings of the 2019 Conference of the North {A}merican
  Chapter of the Association for Computational Linguistics: Human Language
  Technologies, Volume 1 (Long and Short Papers)}, pages 1415--1420,
  Minneapolis, Minnesota. Association for Computational Linguistics.
\newblock

\bibitem[{Zhang et~al.(2019{\natexlab{a}})Zhang, Li, and Song}]{gcn-1.2}
Chen Zhang, Qiuchi Li, and Dawei Song.
\newblock Aspect-based sentiment classification with aspect-specific graph
  convolutional networks.
\newblock In \emph{Proceedings of the 2019 Conference on Empirical Methods in
  Natural Language Processing and the 9th International Joint Conference on
  Natural Language Processing (EMNLP-IJCNLP)}, pages 4568--4578, Hong Kong,
  China. Association for Computational Linguistics.
\newblock

\bibitem[{Zhang et~al.(2019{\natexlab{b}})Zhang, Zhang, Liu, Lin, and
  Xia}]{hd-dataset-1}
Dongyu Zhang, Heting Zhang, Xikai Liu, Hongfei Lin, and Feng Xia.
\newblock Telling the whole story: A manually annotated {C}hinese dataset for
  the analysis of humor in jokes.
\newblock In \emph{Proceedings of the 2019 Conference on Empirical Methods in
  Natural Language Processing and the 9th International Joint Conference on
  Natural Language Processing (EMNLP-IJCNLP)}, pages 6401--6406, Hong Kong,
  China. Association for Computational Linguistics.
\newblock

\bibitem[{Zhang et~al.(2018{\natexlab{a}})Zhang, Wang, and Liu}]{survey-past-2}
Lei Zhang, Shuai Wang, and Bing Liu.
\newblock Deep learning for sentiment analysis: A survey.
\newblock \emph{Wiley Interdisciplinary Reviews: Data Mining and Knowledge
  Discovery}, 8(4):e1253.

\bibitem[{Zhang et~al.(2018{\natexlab{b}})Zhang, Hu, Guo, and Mao}]{cnn-2}
Richong Zhang, Zhiyuan Hu, Hongyu Guo, and Yongyi Mao.
\newblock Syntax encoding with application in authorship attribution.
\newblock In \emph{Proceedings of the 2018 Conference on Empirical Methods in
  Natural Language Processing}, pages 2742--2753, Brussels, Belgium.
  Association for Computational Linguistics.
\newblock

\bibitem[{Zhang and Singh(2018)}]{nlm-1}
Zhe Zhang and Munindar Singh.
\newblock {L}imbic: Author-based sentiment aspect modeling regularized with
  word embeddings and discourse relations.
\newblock In \emph{Proceedings of the 2018 Conference on Empirical Methods in
  Natural Language Processing}, pages 3412--3422, Brussels, Belgium.
  Association for Computational Linguistics.
\newblock

\bibitem[{Zhao et~al.(2019)Zhao, Cattle, Papalexakis, and Ma}]{ngram-4}
Zhenjie Zhao, Andrew Cattle, Evangelos Papalexakis, and Xiaojuan Ma.
\newblock Embedding lexical features via tensor decomposition for small sample
  humor recognition.
\newblock In \emph{Proceedings of the 2019 Conference on Empirical Methods in
  Natural Language Processing and the 9th International Joint Conference on
  Natural Language Processing (EMNLP-IJCNLP)}, pages 6375--6380, Hong Kong,
  China. Association for Computational Linguistics.
\newblock

\bibitem[{Zhong et~al.(2019)Zhong, Wang, and Miao}]{transformer-2}
Peixiang Zhong, Di~Wang, and Chunyan Miao.
\newblock Knowledge-enriched transformer for emotion detection in textual
  conversations.
\newblock In \emph{Proceedings of the 2019 Conference on Empirical Methods in
  Natural Language Processing and the 9th International Joint Conference on
  Natural Language Processing (EMNLP-IJCNLP)}, pages 165--176, Hong Kong,
  China. Association for Computational Linguistics.
\newblock

\bibitem[{Zhou et~al.(2018)Zhou, Yang, and He}]{er-sa}
Deyu Zhou, Yang Yang, and Yulan He.
\newblock Relevant emotion ranking from text constrained with emotion
  relationships.
\newblock In \emph{Proceedings of the 2018 Conference of the North {A}merican
  Chapter of the Association for Computational Linguistics: Human Language
  Technologies, Volume 1 (Long Papers)}, pages 561--571, New Orleans,
  Louisiana. Association for Computational Linguistics.
\newblock

\end{thebibliography}
\bibliographystyle{acl_natbib}

\end{document}